\documentclass{llncs}

\usepackage{amssymb}
\usepackage{amsmath}
\usepackage{leftidx}
\usepackage{graphicx}
\graphicspath{ {Figures/} }

\begin{document}

\title{Motion-Compensated Autonomous Scanning \\for Tumour Localisation using \\Intraoperative Ultrasound}
\titlerunning{Motion-Compensated Autonomous Scanning for Tumour Localisation}  % abbreviated title (for running head)
% \author{Ivar Ekeland\inst{1} \and Roger Temam\inst{2}
% Jeffrey Dean \and David Grove \and Craig Chambers \and Kim~B.~Bruce \and
% Elsa Bertino}
\author{Lin Zhang\inst{1} \and Menglong Ye\inst{1} \and Stamatia Giannarou\inst{1} \and Philip Pratt\inst{2} \and Guang-Zhong Yang\inst{1}}
\authorrunning{L. Zhang et al.} % abbreviated author list (for running head)
\tocauthor{Ivar Ekeland, Roger Temam, Jeffrey Dean, David Grove,
Craig Chambers, Kim B. Bruce, and Elisa Bertino}
% %

\institute{The Hamlyn Centre for Robotic Surgery, Imperial College London, UK\\
\and
Department of Surgery and Cancer, Imperial College London, UK\\
\email{lin.zhang11@imperial.ac.uk}}

\maketitle              % typeset the title of the contribution

\begin{abstract}
Intraoperative ultrasound facilitates localisation of tumour boundaries during minimally invasive procedures. Autonomous ultrasound scanning systems have been recently proposed to improve scanning accuracy and reduce surgeons' cognitive load. However, current methods mainly consider static scanning environments typically with the probe pressing against the tissue surface. In this work, a motion-compensated autonomous ultrasound scanning system using the da Vinci\textsuperscript{\textregistered} Research Kit (dVRK) is proposed. An optimal scanning trajectory is generated considering both the tissue surface shape and the ultrasound transducer dimensions. A robust vision-based approach is proposed to learn the underlying tissue motion characteristics. The learned motion model is then incorporated into the visual servoing framework. The proposed system has been validated with both phantom and \textit{ex vivo} experiments using the ground truth motion data for comparison.

\end{abstract}
\section{Introduction}
In the past decade, intraoperative ultrasound has been increasingly used in both minimally invasive interventions and robot-assisted laparoscopy to assist the localisation and targeting of pathological tissue during tumour dissection. It enables intraoperative visualisation of the tissue without involving ionizing radiation and facilitates the surgeon to define optimal dissection planes. In addition, injury to critical anatomical structures can be avoided, as localisation of margins for tumour dissection is improved. However, manually controlled ultrasound scanning can significantly increase the cognitive load of surgeons, due to the need for maintaining optimal scanning orientation and consistent contact with the tissue, and simultaneously interpreting the ultrasound images during scanning.

Recently, it has been shown that automation of repetitive surgical tasks or tasks that require high precision can be beneficial in improving accuracy and reducing the surgeon's cognitive load. In \cite{mohareri2015us}, an auxiliary robotic arm holding an ultrasound probe has been used to follow the motion of a tele-operated arm such that image guidance can be provided to the surgeon during tumour resection. A 6-DoF ultrasound visual servoing approach has been proposed in \cite{nadeau2013intensity} to control the motion of a robotic system equipped with an ultrasound probe based on intensity information from ultrasound B-mode images. Surgical suturing subtasks have been automated in \cite{shademan2016science} by using a dedicated suturing tool under fluorescence guidance.

An automatic tumour dissection framework using intraoperative ultrasound has been proposed in \cite{pratt2015autous} but it has been designed for planar tissue surfaces. In \cite{zhang2016autonomous}, a framework for autonomous scanning with an endomicroscopy probe over complex tissue surfaces has been proposed to perform microscopic image mosaicking. The above autonomous systems have been designed to deal with static surgical environments which limit their applicability in real surgical scenarios. Hence, a 6-DoF autonomous scanning approach that adapts to tissue motion has yet to be proposed. Tissue motion recovery is important for autonomous execution of surgical tasks. In \cite{mountney2010mcslam}, a respiratory model of tissue motion has been extracted for simultaneously estimating camera motion and tissue deformation. Recently, a probabilistic tracking and surface mapping method was proposed for free-form tissue deformation recovery \cite{giannarou2016deform}.
 
In this paper, a novel system for autonomous ultrasound scanning under tissue motion is presented for intraoperative tumour localisation. An efficient approach is proposed to plan in advance the optimal trajectory for the ultrasound scanning, considering the surface curvature as well as ultrasound transducer dimensions for efficient coverage, which optimises the ultrasound image quality and scanning time. For motion compensation, a robust vision-based approach is proposed to learn the tissue motion. The learned motion model is updated online and incorporated into the scanning movement of the ultrasound probe. A 3D tumour model is reconstructed from a sequence of ultrasound images obtained after scanning, which provides an intuitive visualisation of tumour location under the tissue surface. The proposed system has been validated on phantom and \textit{ex vivo} experiments using ground truth motion data and CT ground truth. The presented results verify that our system allows the Field-of-View (FoV) of the ultrasound to follow the tissue motion, which simplifies the tumour localisation task for the surgeon.

\section{Methods}
\begin{figure}[t]
	\centering
	\includegraphics[width=\linewidth]{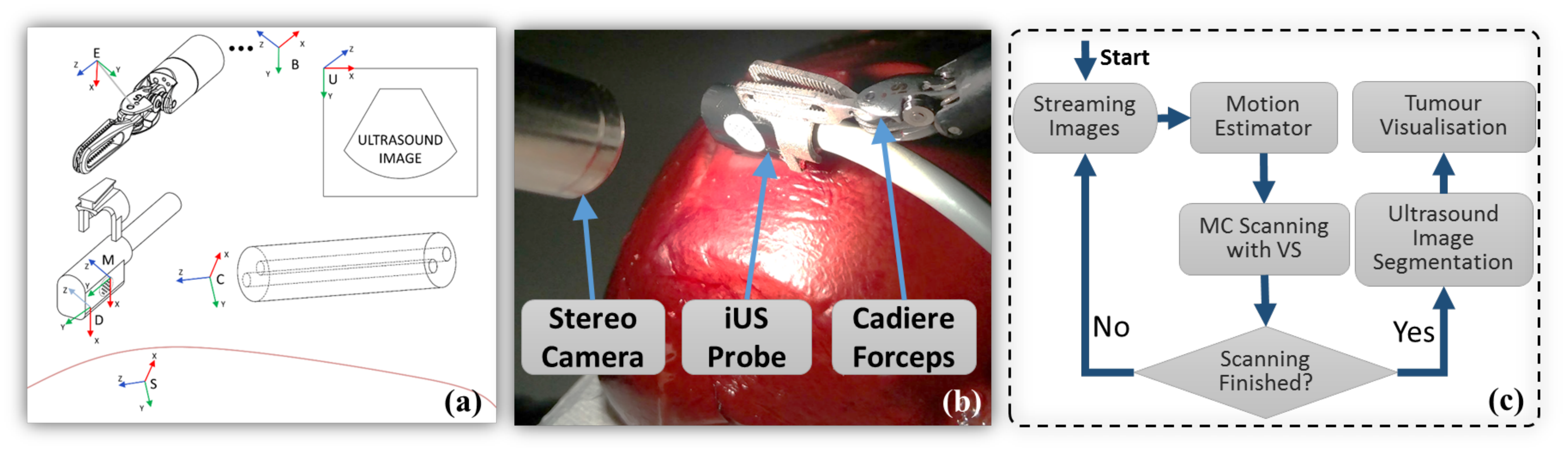}
	\caption{(a) Coordinate frames defined in the proposed system. B: robot base; E: robot end-effector; M: marker attached on the probe; D: transducer; C: stereo camera; S: tissue surface; (b) Hardware Setup; (C) Flowchart of the proposed framework. MC: Motion Compensated; VS: Visual Servoing.}
	\label{fig:system}
\end{figure}

\subsection{Motion Estimation and Online Update}
For motion compensation, the tissue motion can be learned prior to scanning, before the ultrasound probe entering the FoV of the stereo camera, thereby eliminating occlusion of the scene. At time $t$, a 3D surface point $ \mathbf{P}_t $ is estimated by matching its 2D projections $\mathbf{p}_{t}^{l}$ and $ \mathbf{p}_{t}^{r} $ on the left and right images of the camera, respectively, using the stereo matching method proposed in \cite{yang2010csbp}.

To recover the global motion of the tissue, the relative motion $ \Delta \mathbf{P}_{t} = \mathbf{P}_{0} - \mathbf{P}_{t}$ of a set of points on the tissue surface is calculated. The first stereo image pair at time $t=0$ is denoted as the reference and all the following images are compared to it. We estimate the optical flow \cite{farneback2003flow} between the reference image and the image at time $t$ to find the correspondence pair $ \left\lbrace\mathbf{p}_{0}^{l}, \mathbf{p}_{t}^{l} \right\rbrace$. For accurate motion estimation, a grid of points are sampled from the target scanning region, which are tracked as above. To eliminate outlier correspondences, forward-and-backward tracking is performed and the inconsistent points are discarded in the motion estimation. 
The tissue motion $ \Delta \mathbf{P}_{t}$ is estimated as the median of the motion of the consistent grid points. 

In order to learn the underlying motion characteristics, Principle Component Analysis (PCA) is used to extract the major motion and project it from the camera coordinate space to the PCA space. The PCA component with the largest eigenvalue corresponds to the respiratory motion.

In this paper, a typical respiratory model with asymmetric period similar to \cite{mountney2010mcslam} is employed. The model can be represented as:
\begin{equation}\label{eq:res_model}
	z(t) = z_{0} - b \cos^{2n}(\frac{\pi t}{\tau}-\phi)
\end{equation}
where $z_0$ is the exhale position of the respiration, $b$ is the amplitude, $\tau$ is the period of the respiration, $\phi$ is the phase and $n$ describes the degree of asymmetry (the larger the $n$, the longer the expiration). In this work, $n=3$ which represents a typical respiratory motion model \cite{lujan2003respiration}. The parameters of the model are estimated such that the model fits to the major motion in the PCA space. This is treated as a non-linear least-squares problem which can be solved by the Levenberg-Marquardt optimisation method. Unlike \cite{mountney2010mcslam}, where a fixed phase $\phi=0$ has been assumed, all four parameters ($z_0$, $b$, $\tau$ and $\phi$) are estimated in this paper. 

\subsubsection{Online Motion Update} Small errors in the period of the learned motion ($\tau$) can result in the latency between the robot and tissue motion during scanning. In particular, the phase of the learned model will be gradually shift away from the actual tissue motion. To correct this phase shift, the tissue motion model is also estimated online during ultrasound scanning. During scanning, the user can define a region on the tissue for online motion estimation. We use the same 3D reconstruction result from \cite{yang2010csbp}, and perform optical flow tracking on the selected tissue region. With the tracked 2D points, tissue movements can be estimated by solving a Perspective-n-Point (PnP) problem. Online motion estimation is performed using the tissue movements in the last two cycles, and updates only the parameter $\hat{\phi}$ during scanning, as we found phase is an effective parameter for adjusting the latency between the robot and tissue motion.

\subsection{Motion-Compensated Scanning}
To generate a desired trajectory for ultrasound scanning, a target region on a reference image is defined by the user. A starting point is selected and a zig-zag trajectory which covers the required area is planned on the 2D reference image. Benefiting from stereo matching, the corresponding 3D positions of these 2D trajectory points can be acquired. The distance between each line of the zig-zag trajectory is defined as half width of the transducer in order to scan the region efficiently. At each 3D point along the trajectory, surface normal is calculated using its neighbouring points. The coordinate systems are defined in Fig. \ref{fig:system}(a). The scanning trajectory is planned with respect to the pose of the marker (circular-dot pattern) attached on the ultrasound transducer. It is defined as a sequence of desired marker poses ($\mathbf{T}_{C}^{M*}$) in the camera coordinate system such that the transducer can be aligned with the surface normal at every point of the trajectory.

In this work, a position-based visual servoing method is employed to control the robot in the Cartesian space.
More specifically, a desired pose for the robot end-effector is calculated in each control loop as:
\begin{equation}\label{eq:end_pose}
	\mathbf{T}_{B}^{E*} = \mathbf{T}_{B}^{E} \cdot \mathbf{T}_{E}^{M} \cdot \mathbf{\widetilde{T}}_{M}^{M*} \cdot \mathbf{\hat{T}}_{M}^{M*} \cdot \mathbf{T}_{M*}^{E*},
\end{equation}
where $\mathbf{T}_{B}^{E}$ is the current end-effector pose in the robot base coordinate system. $\mathbf{T}_{E}^{M} = (\mathbf{T}_{M*}^{E*})^{-1}$ is a constant transformation between the end-effector and the marker on the probe that can be measured. The desired relative movement of the marker depends on the motion caused by the scanning motion as well as the motion for compensating for respiration, which are denoted as $\mathbf{\widetilde{T}}_{M}^{M*}$ and $\mathbf{\hat{T}}_{M}^{M*}$, respectively.
The desired relative marker pose for scanning is calculated as:
\begin{equation}\label{eq:scan_mHmd}
	\mathbf{\widetilde{T}}_{M}^{M*} = \mathbf{\hat{T}}_{M}^{M*} \cdot (\mathbf{T}_{C}^{M})^{-1} \cdot \mathbf{\widetilde{T}}_{C}^{M*}.
\end{equation}
Here, $ \mathbf{\widetilde{T}}_{C}^{M*} $ and $\mathbf{T}_{C}^{M}$ are the desired marker pose and the detected marker pose (in the camera coordinate system) in the reference and current frame, respectively.
The desired relative marker pose for motion compensation is calculated as:
\begin{equation}\label{eq:motion_mHmd}
	\mathbf{\hat{T}}_{M}^{M*} = \mathbf{T}_{M}^{D} \cdot \mathbf{T}_{D}^{S} \cdot \mathbf{T}_{S}^{S*} \cdot \mathbf{T}_{S*}^{D*} \cdot \mathbf{T}_{D*}^{M*},
\end{equation}
where $\mathbf{T}_{M}^{D}=\mathbf{T}_{M*}^{D*}$ is a constant offset from the marker to the transducer that can be measured. $\mathbf{T}_{D}^{S}=\mathbf{T}_{S*}^{D*}$ is a rotation between the transducer and surface point which has the same orientation as the camera coordinate. 
$ \mathbf{T}_{S}^{S*} = \left( \begin{smallmatrix} \mathbf{I}_3 & \Delta \mathbf{x}\\ 0&1 \end{smallmatrix} \right) $ is the transformation from the current to the next surface position which can be calculated from the learned motion model using the PCA transformation. In particular, the tissue position $\mathbf{x}_{t}$ relative to the reference frame at time $t$ can be estimated via: $\mathbf{x}_{t} = \mathbf{w}^{-1} \cdot [z(t),0,0]^\intercal$, in which $\mathbf{w}$ is the PCA eigenvectors describing the transformation from 3D space to the PCA space. Then $\Delta \mathbf{x} = \mathbf{x}_{t+1} - \mathbf{x}_{t}$.

\subsection{Tumour Segmentation and Visualisation}
To create a 3D mesh of the tumour for intraoperative visualisation, the ultrasound images captured during scanning are segmented. Any segmentation method can be integrated into the system (manual in this paper). The tumour segmented boundary is represented as a number of points where each point is connected to its neighbours across ultrasound images so that a 3D mesh can be reconstructed. The segmented boundary points are transformed from the 2D image coordinate system to the 3D camera coordinate system {C}. To this end, an ultrasound coordinate system {U} is defined in which each pixel of the ultrasound image can be represented as a 3D point $P^U$. The segmented boundary points are then transformed to the coordinate {C} via: $P^C=\mathbf{T}_{M}^{C} \cdot \mathbf{T}_{D}^{M} \cdot \mathbf{T}_{U}^{D} \cdot P^U$ where $\mathbf{T}_{U}^{D}$ can be measured. The created tumour mesh is registered in {C} as shown in Fig. \ref{fig:scan_demo}(d) and (h).
\begin{table}[t]
\scriptsize
\renewcommand{\arraystretch}{1.2}
\centering
\caption{Estimation results of parameter on three different motion profiles}
\label{table:mot_est}
\begin{tabular}{ccccccc}
\hline\noalign{\smallskip}
                     & \multicolumn{2}{c}{Profile 1} & \multicolumn{2}{c}{Profile 2} & \multicolumn{2}{c}{Profile 3} \\
                     & $\tau$ (\textbf{Frames})	& $b$ (\textbf{mm})	& $\tau$ (\textbf{Frames})	& $b$ (\textbf{mm})	& $\tau$ (\textbf{Frames})	& $b$ (\textbf{mm})       \\
\noalign{\smallskip}\hline
Phantom Estimated    &	75.20$\pm$0.25	&	2.98$\pm$0.08	&	125.10$\pm$0.39	&	3.02$\pm$0.10	&	124.93$\pm$0.99	&	4.96$\pm$0.12   \\
Phantom Ground Truth &	75.00	&	3.00	&	125.00	&	3.00	&	125.00	&	5.00  \\
Ex Vivo Estimated    &	75.08$\pm$0.22	&	3.16$\pm$0.16	&	125.15$\pm$0.28	&	3.08$\pm$0.28	&	125.10$\pm$0.41	&	4.82$\pm$0.36  \\
Ex Vivo Ground Truth &	75.00	&	3.00	&	125.00	&	3.00	&	125.00	&	5.00   \\ \hline      
\end{tabular}
\end{table}
\section{Results}
\noindent\textbf{Experimental Setup} The proposed autonomous ultrasound scanning system includes an Aloka ultrasound machine with an UST-533 linear transducer, a standard da Vinci\textsuperscript{\textregistered} surgical robot, and a workstation for real-time processing and robot control. The robot includes a sterescopic camera streaming images at 25Hz, and a patient site manipulator which is controlled via the dVRK. As shown in Fig.\ref{fig:system}(b), a circular-dot pattern (KeySurgical Inc., USA) has been attached to the ultrasound probe to facilitate tool tracking and pose estimation \cite{pratt2015autous} at 25Hz. The motion estimation module (including optical flow and parameter estimation) runs at $\approx$30Hz. To evaluate our proposed framework, experiments were performed on two kidney phantoms (made of polyvinyl alcohol) and an \textit{ex vivo} tissue (ovine liver). For \textit{ex vivo} experiments, tumours are simulated using olives which are embedded inside the tissue.
\begin{figure}[t]
	\centering
	\includegraphics[width=\linewidth]{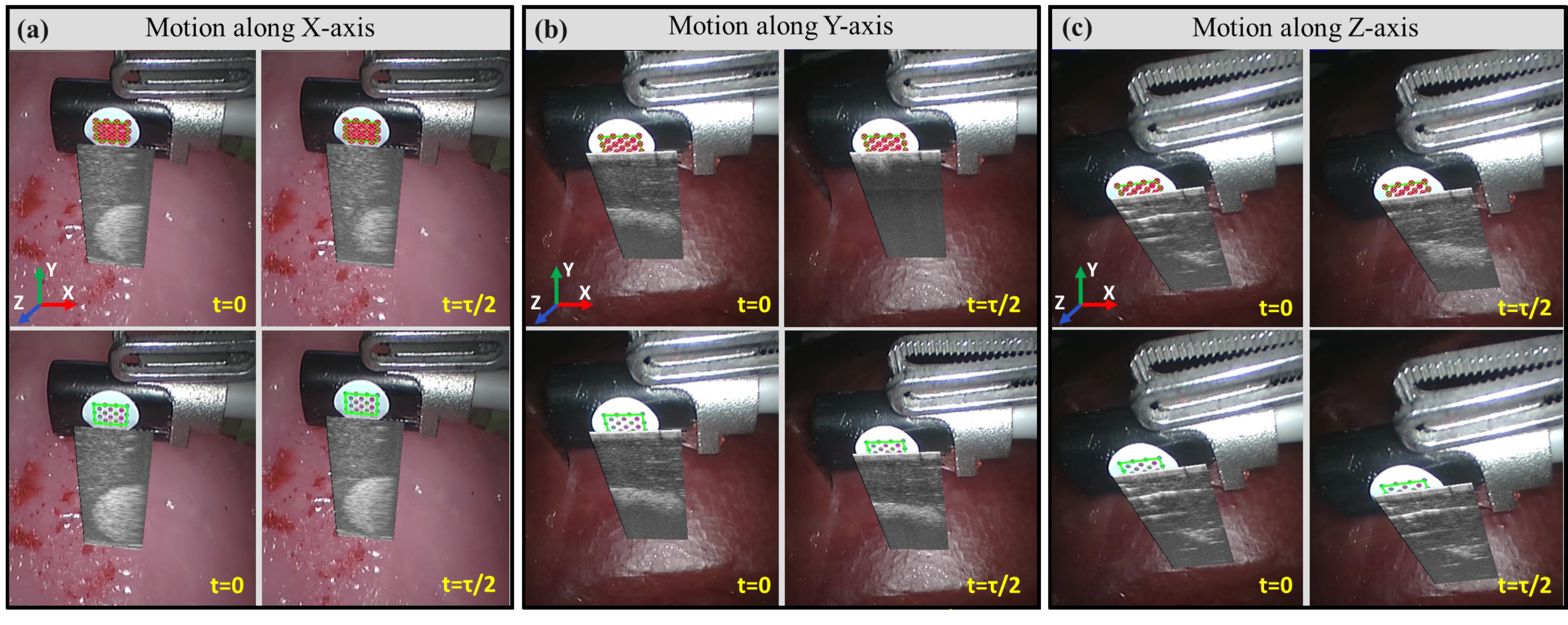}
	\caption{Qualitative results of ultrasound stabilisation. The top and bottom rows are the trials without and with motion compensation, respectively. Three motion profiles were tested on a phantom (a) and \textit{ex vivo} tissue (b-c), with directions along the X-, Y- and Z-axis of the motorised device.}
	\label{fig:us_comp_demo}
\end{figure}

\begin{figure}[t]
	\centering
	\includegraphics[width=\linewidth]{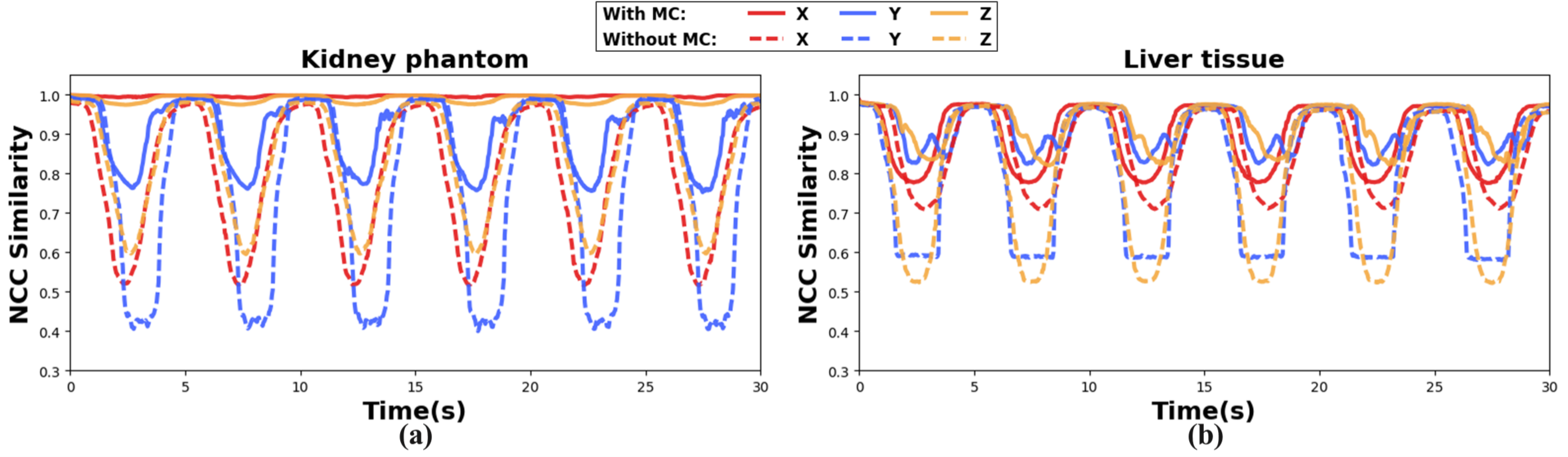}
	\caption{Quantitative results of ultrasound stabilisation on phantoms (a) and \textit{ex vivo} tissue (b) with and without motion compensation (MC). The accuracy is represented using the NCC similarity between each image to the reference image.}
	\label{fig:us_comp_similarity}\vspace*{-\baselineskip}
\end{figure}
\noindent\textbf{Motion Estimation and Stabilisation} To simulate respiratory motion, a motorised device is used to host the targets for scanning, such that periodic motion can be generated. Both phantom and \textit{ex vivo} tissue are moved by the device, and three motion profiles with different periods ($\tau \in \lbrace 75, 125 \rbrace$) and amplitudes ($b \in \lbrace 3, 5 \rbrace $) are simulated. For each motion profile, we conduct 9 trials to estimate motion parameters $\tau$ and $b$. These trials contain 3 trials each for motion along $X$, $Y$, and $Z$ axis (w.r.t. motorised device coordinate system). Table \ref{table:mot_est} presents the mean and standard deviation values of these parameters. It can be seen that our motion estimation provides accurate results compared to the ground truth models (defined on the motorised device). 

To further demonstrate the influence of these accuracies, we apply motion compensation for ``ultrasound stabilisation''. As shown in Fig.\ref{fig:us_comp_demo}, the robot is controlled to adapt to certain surface points on the tissue in the presence of tissue motion. The difference in obtained ultrasound images show the effect of motion estimation. The stabilisation accuracy is quantified using normalised cross-correlation (NCC). The NCC similarity (range of $\left[ 0,1\right]$) is calculated between each image and the reference image obtained at exhale position. A high NCC measure indicates a good quality of motion estimation and compensation. Fig.\ref{fig:us_comp_similarity} presents the NCC measures calculated with and without motion compensation on three different motion profiles. It is clear that by introducing motion estimation and compensation, the NCC measures can be significantly improved.
\begin{figure}[t]
	\centering
	\includegraphics[width=\linewidth]{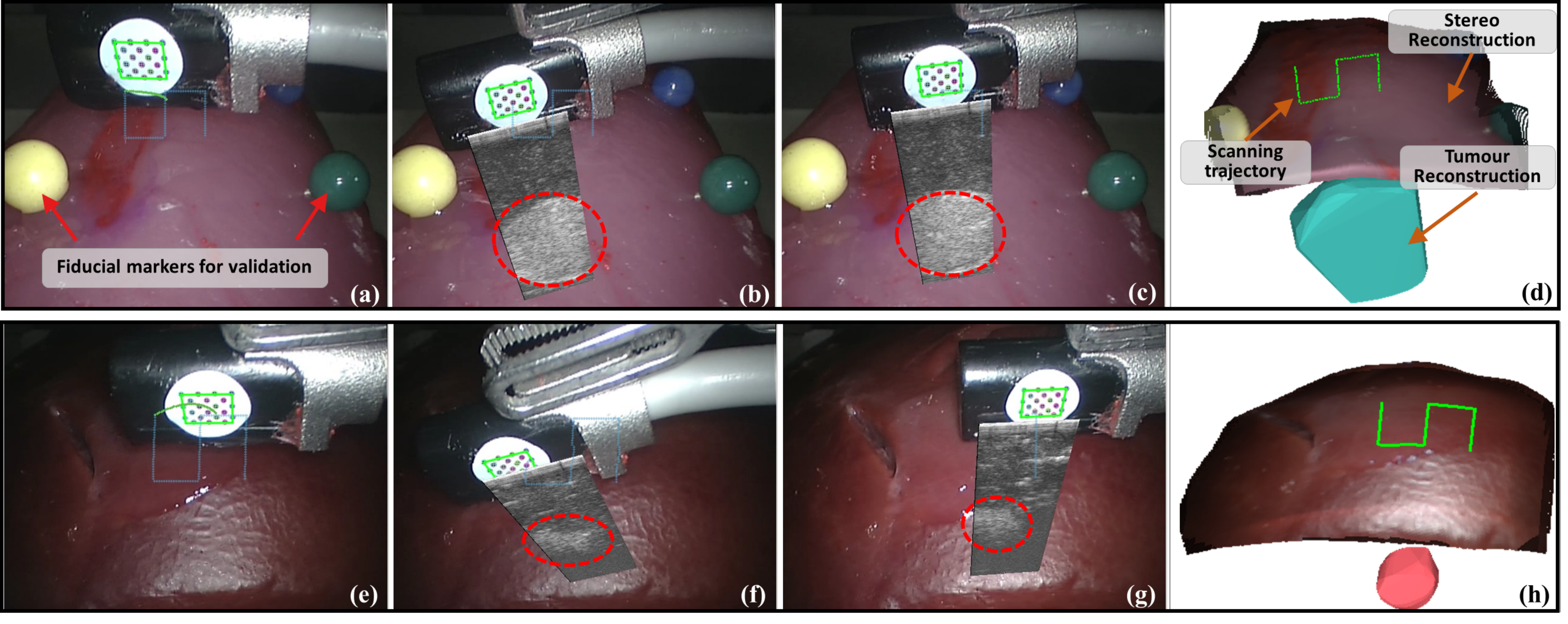}
	\caption{Qualitative results of two example autonomous scanning tasks on a phantom (a-d) and \textit{ex vivo} tissue (e-h). The obtained tumours and 3D tissue surfaces are visualised in (d) and (h). See supplementary videos for details.}
	\label{fig:scan_demo}
\end{figure}
\begin{table}[t]
\scriptsize
\renewcommand{\arraystretch}{1.2}
\centering
\caption{Quantitative results of tumour scanning under three different motion profiles. Tumour localisation errors are indicated using the differences in location and size (diameter) between obtained tumours and CT ground truth. }
\label{table:tumour_error}
\begin{tabular}{ccccccc}
\hline\noalign{\smallskip}
                     & \multicolumn{3}{c}{Phantom 1} & \multicolumn{3}{c}{Phantom 2} \\
                     & Profile 1	& Profile 2	& Profile 3	& Profile 1	& Profile 2	& Profile 3\\
\noalign{\smallskip}\hline
Location Error (mm)    &	3.85	&	2.33	&	3.81	&	2.78	&	2.57	&	3.38   \\
Diameter Error (mm) &	0.69	&	0.61	&	0.33	&	0.84	&	0.61	&	0.77   \\ \hline   
\end{tabular}
\end{table}

\noindent\textbf{Tumour Scanning under Tissue Motion} To evaluate the performance of tumour scanning under tissue motion, further experiments are conducted. For a scanning target (phantom/tissue), a user defines a point-of-interest for scanning and a region for motion estimation. Our framework then performs autonomous scanning with motion compensation, followed by ultrasound image segmentation and tumour visualisation. The qualitative results are presented in Fig.\ref{fig:scan_demo}. To validate the reliability of scanning, the locations and sizes of the obtained tumours are compared to the ground truth (via CT scanning). As shown in Table \ref{table:tumour_error}, the proposed framework presents accuracy results amounting to 0.64 mm (size) and 3.12 mm (location), which are close to the clinically accepted tumour dissection margin errors (5-7 mm) reported in \cite{hughes2014ar}. It is worth noting that, the location errors include the registration error between the CT and camera coordinate systems. Supplementary video is available via: \url{https://youtu.be/HszuMXFXqmU}.

\section{Conclusions}
In this paper, we presented a motion-compensated autonomous system for intraoperative ultrasound scanning during robotic surgery. The proposed system performs vision-based tissue motion estimation with online update to recover periodic respiratory motion. The estimated motion parameters are then combined with 6-DOF visual servoing for accurate continuous scanning under tissue motion. The obtained ultrasound images are segmented such that tumours can be localised and visualised to the surgeons. Both the quantitative and qualitative results from phantom and \textit{ex vivo} experiments demonstrate the capability of our system for motion-compensated autonomous scanning. The proposed system allows the FoV of the ultrasound to follow the tissue motion, which not only simplifies scanning tasks for surgeons but can also permit observation of temporal changes of the underlying tissue such as blood flow and contrast uptake.
\bibliographystyle{splncs03}
\bibliography{mybib}

\end{document}